\documentclass[runningheads]{llncs}
\usepackage[T1]{fontenc}
\usepackage{graphicx,verbatim}
\usepackage{subcaption}
\usepackage{svg}
\usepackage{multirow}
\usepackage{dsfont}
\usepackage{graphicx}
\usepackage{comment}
\usepackage{amsmath}
\usepackage{booktabs}
\usepackage{newclude}
\usepackage{amsfonts,amssymb}
\usepackage[hidelinks]{hyperref} 
\usepackage{xspace}
\usepackage{colortbl}
\usepackage[dvipsnames,svgnames]{xcolor}
\definecolor{Ash grey}{rgb}{0.7, 0.75, 0.71}
\newcommand{\model}{{HyperVLP}\xspace}

\newcommand{\eat}[1]{}

\usepackage{marvosym}
\begin{document}
\title{HyperVLP: Enhancing Hierarchical Surgical Video-Language Pre-training in Hyperbolic Space}

\author{Yaojun Hu\inst{1} \and
Kun Yuan\inst{2,3,4}$^\dagger$ \and 
Nassir Navab\inst{4} \and \\
Haochao Ying\inst{1}$^\dagger$ \and
Jian Wu\inst{1} \and
Nicolas Padoy\inst{2,3}
}
\institute{Zhejiang University, Hangzhou, China \and
University of Strasbourg, CNRS, INSERM, ICube, UMR7357, Strasbourg, France \and
IHU Strasbourg, Strasbourg, France \and
Technical University of Munich, Munich, Germany \\
\email{yaojunhu@zju.edu.cn, kyuan@unistra.fr, haochaoying@zju.edu.cn} 
}
\footnotetext[1]{Corresponding authors: Kun Yuan and Haochao Ying.}
\footnotetext[2]{This work has been accepted to MICCAI 2026.}
\maketitle              
\begin{abstract}
Surgical vision-language foundation models typically adopt educational materials, such as surgical lecture videos, to transfer surgical knowledge encoded in language into visual representations. These knowledge are multi-dimensional and hierarchical: fine-grained action cues appear in narration, mid-level key steps are summarized in subsection headings, and global procedural context, such as patient history and surgical strategy, is described in abstract texts. Prior work largely collapses these heterogeneous signals into a single flat embedding space, implicitly assuming independence across hierarchy levels. However, this is suboptimal because it ignores cross-level semantic containment, e.g., actions belong to steps, steps compose phases, weakens long-range dependency modeling. 
To this end, we propose a hyperbolic surgical video-language pre-training framework that explicitly preserves the hierarchical structure by mitigating structural false negatives induced by procedural context and enforcing semantic consistency between parent phases and their constituent child steps.  
Extensive experiments on multiple surgical benchmarks show consistent gains in zero- and few-shot phase recognition across procedures and institutions. 


\end{abstract}

\section{Introduction}
Surgical workflow analysis aims to automatically interpret intraoperative activities from video, with surgical phase recognition as a core task to model procedural progression and provide support for clinical decision-making and objective surgical assessment~\cite{maier2022surgical_background, liu2025survey}. 
Although supervised methods achieve strong performance~\cite{Pér_MuST_MICCAI2024, Yan_Surgformer_MICCAI2024}, they depend on dense annotations and generalize poorly across institutions and unseen procedures. 
Vision-language pretraining (VLP)~\cite{radford2021clip, li2022blip, wang2022medclip, pmc-clip} has emerged as a powerful paradigm for zero-/few-shot generalization, which allows models to transfer knowledge from the educational corpora to new tasks by aligning visual and textual representations. 

Several important contributions in vision-language pre-training have been made in surgical workflow analysis~\cite{schmidgall2024general, wang2023foundation, he2025endo}. SurgVLP~\cite{yuan2023SurgVLP} pioneers CLIP-style pretraining for surgical workflows with large-scale clip-caption datasets spanning multiple surgical types, while HecVL~\cite{Yuan2024HecVL} and PeskaVLP~\cite{peskavlp} further introduce hierarchical supervision in euclidean embedding space and LLM-based language refinement to enrich structured supervision. Related efforts such as SurgLaVi~\cite{perez2025surglavi} further constructs a hierarchical large-scale surgical video-language dataset to support domain-specific pretraining. 

Surgical workflows exhibit a hierarchical tree structure~\cite{perez2025surglavi, peskavlp, Yuan2024HecVL}, where phases decompose into ordered steps and fine-grained actions. 
Existing methods embed these relations in Euclidean space, which lacks the geometric inductive bias to preserve cross-level entailment, weakening parent-child dependencies and limiting structured reasoning and cross-procedure generalization. 
In contrast, hyperbolic space~\cite{cannon1997hyperbolicgeometry,Desai2023meru,PalSDFGM2024hypercoclip}, with negative curvature and exponential capacity, naturally supports low-distortion embedding of hierarchical data and semantic containment. 
Moreover, current contrastive objectives~\cite{yuan2023SurgVLP, perez2025surglavi, radford2021clip} treat all non-matching video-text pairs as negatives, even when segments are semantically adjacent or temporally correlated. Such design ignores the inherent tree structure and pushes clinically related nodes apart, introducing structural false negatives and distorting workflow hierarchical tree structure. 

To resolve the mismatch between tree-structured surgical knowledge and flat Euclidean embeddings, we propose \model, a hyperbolic video-language pre-training framework. We project video and text representations into a shared Lorentz hyperbolic manifold that naturally models hierarchical containment. To mitigate structural false negatives introduced by conventional contrastive learning, we design a geometry-aware hyperbolic contrastive loss that down-weights semantically related samples within the same procedural context. To further enforce clinically meaningful parent-child dependencies, we introduce a cone-based hyperbolic entailment objective that imposes geometric containment between parent and child embeddings across both inter-modal and intra-modal relations. Since hierarchical constraints in hyperbolic space can destabilize early alignment~\cite{ganea2018hyperboliccones,le-etal-2019-inferring}, we adopt a progressive two-stage optimization strategy that first learns cross-modal alignment via geometry-aware contrastive learning and then enforces hierarchical containment through entailment-based refinement. We evaluate \model on Cholec80~\cite{twinanda2016endonetcholec80}, AutoLaparo~\cite{wang2022autolaparo}, and MultiBypass140~\cite{lavanchy2024challengesmultibypass}, demonstrating consistent gains in phase recognition under both zero- and few-shot settings. 
Overall, our contributions are summarized as follows:
\begin{itemize}
\item We propose \model, a hierarchical surgical video-language pretraining framework in hyperbolic space that explicitly encodes the intrinsic tree structure of surgical workflows.

\item We identify structural false negatives in surgical video-language contrastive learning and develop a geometry-aware hyperbolic contrastive strategy for hierarchy-aware alignment.

\item We introduce a cone-based hyperbolic entailment objective that enforces geometric containment between parent and child embeddings across both inter- and intra-modal relations.

\end{itemize}

\section{Method}

\begin{figure}[t]
    \centering
    \includegraphics[width=0.99\linewidth]{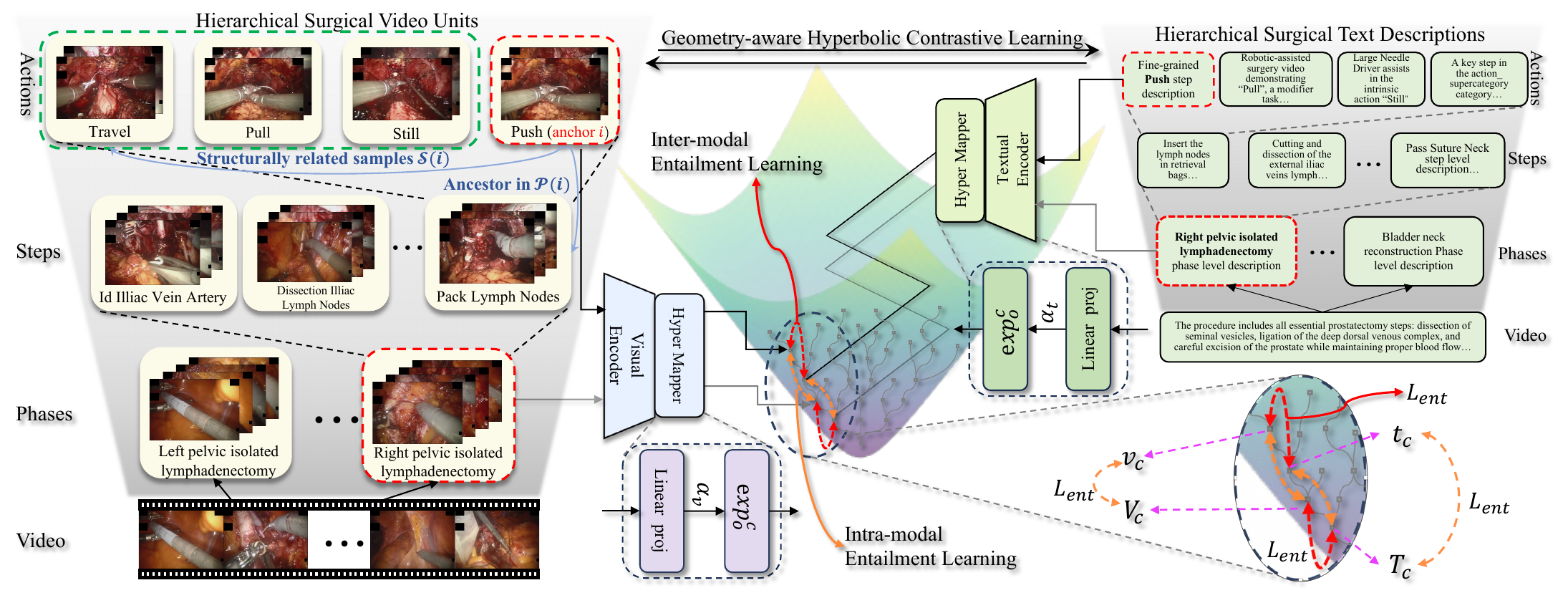}
    \caption{
Overview of HyperVLP. Red dashed boxes denote hierarchical units $(V,T,v,t)$, which are mapped to hyperbolic embeddings $(V_c, T_c, v_c, t_c)$ shown on the right. Green boxes indicate structurally related samples $\mathcal{S}(i)$ used in geometry-aware contrastive learning.
}
    \label{fig:main}
\end{figure}

\subsection{Preliminaries: Hyperbolic Geometry}~\label{sec:back}
Hyperbolic space naturally models hierarchical data due to its constant negative curvature and exponential volume growth. This geometry enables low-distortion embedding of tree-structured surgical workflows, where phases decompose into steps and fine-grained actions. We adopt the Lorentz model $\mathbb{L}^n$, defined as the upper sheet of a two-sheeted hyperboloid $\mathbb{L}^n = \{ \mathbf{x} \in \mathbb{R}^{n+1} : \langle \mathbf{x}, \mathbf{x} \rangle_{\mathbb{L}} = - \tfrac{1}{{c}}, 
    x_0 = \sqrt{ \tfrac{1}{{c}} + \| \mathbf{x}_n \|^2 }, \, {c} > 0 \},$ 
where ${c} \in \mathbb{R}$ represents the curvature and $\langle \mathbf{x}, \mathbf{y} \rangle_{\mathbb{L}} = \langle \mathbf{x}_n, \mathbf{y}_n\rangle_{\mathbb{E}} - x_0 y_0$ denotes the Lorentzian inner product with $\langle \cdot, \cdot \rangle_{\mathbb{E}}$ indicating the Euclidean inner product. $\mathbf{x}_n$ is the $n$-dimensional spatial component of $\mathbf{x}$.
The geodesic distance is
$
d_{\mathbb{L}}(\mathbf{x}, \mathbf{y})
= \sqrt{\tfrac{1}{c}} \cosh^{-1}\left(-c\langle \mathbf{x}, \mathbf{y} \rangle_{\mathbb{L}}\right).
$
Euclidean features are projected to $\mathbb{L}^n$ by mapping tangent vectors at the origin $\mathbf{o}=[\sqrt{1/c},\mathbf{0}]$ through the exponential map
$
\exp_{\mathbf{o}}^{c}(\mathbf{x}) 
= \cosh \left( \sqrt{c}\, \| \mathbf{x}\|_{\mathbb{L}} \right) \mathbf{o} 
+ \frac{\sinh \left( \sqrt{c}\, \| \mathbf{x}\|_{\mathbb{L}} \right)}
{\sqrt{c}\, \| \mathbf{x}\|_{\mathbb{L}} } \mathbf{x}
$.

\subsection{Proposed Methodology: HyperVLP}
\noindent \textbf{Hierarchical Embedding in Hyperbolic Space.}
Our data contains three hierarchical levels (e.g., phase, step, action). During training, we sample any two adjacent levels to form a parent-child pair for hierarchical modeling. For each unit $(V,T,v,t)$, visual and textual features are extracted by encoders $\mathcal{F}_v$ and $\mathcal{F}_t$, linearly projected to the tangent space at the origin $\mathbf{o}=[\sqrt{1/c},\mathbf{0}]$, and mapped to the Lorentz manifold $\mathbb{L}^n$ via the exponential map: 
$V_c = \exp_{\mathbf{o}}^{c}(W_v \mathcal{F}_v(V)), \quad
T_c = \exp_{\mathbf{o}}^{c}(W_t \mathcal{F}_t(T)).$
The same transformation is applied to the step-level features $v$ and $t$ to obtain $v_c$ and $t_c$. Learnable scaling in the tangent space stabilizes optimization under negative curvature. 
\\
\\
\noindent \textbf{Geometry-Aware Hyperbolic Contrastive learning.}~\label{sec:cont}
Standard contrastive objectives~\cite{oord2018cpc,radford2021clip, wang2022medclip} always treat all non-matching samples as true negatives. This assumption breaks in hierarchical surgical workflows. We identify two sources of \emph{structural false negatives}:
(i) \textbf{contextual proximity}, where clips correspond to adjacent steps within the same phase;
(ii) \textbf{hierarchical entailment}, where fine-grained clips are semantically subsumed by higher-level segments.
Treating such samples as independent negatives enforces artificial separation and distorts the workflow topology. 

For each anchor $i$, we partition candidates into three sets (illustrated on the left side of Fig.~\ref{fig:main}):
(i) $\mathcal{S}(i)$: structurally related samples from the same procedure grouped by temporal containment (including contextual neighbors); 
(ii)$\mathcal{P}(i)$: matched positive pairs (including hierarchical ancestors and descendants); 
(iii) $\mathcal{N}(i)$: remaining cross-procedure samples. We define the geometry-aware 
hyperbolic contrastive loss between parent-level visual and textual embeddings as
$$
\mathcal{L}_{geohc}(\mathbf{V_c}, \mathbf{T_c})=-\frac{1}{B}
\sum_{i=1}^{B}\log
\frac{
\sum_{k \in \mathcal{P}(i)} {d_{ik}}
}{
\sum_{k \in \mathcal{P}(i)} {d_{ik}} +
\sum_{k \in \mathcal{S}(i)} {\alpha_{ik}d_{ik}}
+
\sum_{k \in \mathcal{N}(i)} {d_{ik}}
}
$$
where $d_{ik} = - d_{\mathbb{L}}(V_c^i, T_c^k)/\tau$ denotes the hyperbolic similarity between anchor $i$ and sample $k$. 
To mitigate structural false negatives, we introduce an adaptive weight
$
\alpha_{ik} = \sigma\!\left(
- \tfrac{d_{ik}}
{\sum_{k \in \mathcal{P}(i)} d_{ik} + \epsilon}
\right)
$
where $\sigma(\cdot)$ is the sigmoid function and $\epsilon$ ensures numerical stability. This design suppresses structurally related samples that exhibit high similarity relative to true positives, thereby preventing over-separation of semantically correlated clips and preserving hierarchical topology. 
We further construct the symmetric formulation by anchoring textual embeddings against visual candidates, resulting in $\mathcal{L}_{geohc}(\mathbf{T}_c,\mathbf{V}_c)$. The same geometry-aware mechanism is applied at the child level to enforce hierarchical generalization. The overall hyperbolic contrastive objective is defined as: 
$$
\mathcal{L}_{GeoHCL}
=
\frac{1}{4}
(
\mathcal{L}_{geohc}(\mathbf{V}_c,\mathbf{T}_c)
+
\mathcal{L}_{geohc}(\mathbf{T}_c,\mathbf{V}_c)
+
\mathcal{L}_{geohc}(\mathbf{v}_c,\mathbf{t}_c)
+
\mathcal{L}_{geohc}(\mathbf{t}_c,\mathbf{v}_c)
).
$$
\noindent \textbf{Hierarchical Entailment Learning.}\label{sec:ent}
Some prior studies~\cite{ganea2018hyperboliccones} investigate hierarchical representation learning in hyperbolic space. In our work, we introduce a hierarchical-aware hyperbolic entailment mechanism that explicitly models multi-level semantic containment across modalities.
Specifically, we adopt hyperbolic entailment cones in the Lorentz space, where each point $\mathbf{x}$ defines a conical region $\mathbb{R}_{\mathbf{x}}$ such that $\mathbf{y} \in \mathbb{R}_{\mathbf{x}}$ is semantically subsumed by $\mathbf{x}$. The half-aperture is given by $\theta(\mathbf{x}) = \sin^{-1}\left(\tfrac{2K}{\sqrt{c}\| \mathbf{x}_{n} \|}\right)$, where $c$ denotes curvature and $K$ controls behavior near the origin. The aperture shrinks with increasing $\| \mathbf{x}_{n} \|$, naturally enforcing tighter semantic specificity away from the origin. 
During training, we constrain a specific concept $\mathbf{y}$ to lie within the aperture $\theta(\mathbf{x})$ of its parent concept $\mathbf{x}$ by penalizing the angular residual of an outward point $\mathbf{y}$ with exterior angle denoted as 
$\varphi(\mathbf{y},\mathbf{x}) =\tfrac{y_0 + c x_0 \langle \mathbf{y},\mathbf{x} \rangle_{\mathcal{L}}}
{\|\tilde{x}\|\sqrt{c\langle \mathbf{y},\mathbf{x} \rangle_{\mathcal{L}}^2 - 1}}$, 
where $y_0$ and $x_0$ denoting the time dimension. Thus, the entailment loss is formulated as: $    L_{ent}(\mathbf{y},\mathbf{x}) = \max \left( 0, \varphi(\mathbf{y},\mathbf{x}) - \eta \theta(\mathbf{x}) \right)$, where $\eta$ is a threshold to adjust the distance region.

To explicitly model semantic containment in surgical workflows, we introduce both inter-modal and intra-modal entailment supervision in hyperbolic space. Across modalities, we treat textual concepts as semantically containing their corresponding videos (e.g., ``cut'' entails all cutting clips), and enforce containment at both child and parent levels via $L_{ent}(v_{c}, t_{c})$ and $L_{ent}(V_{c}, T_{c})$, yielding the inter-modal loss $\mathcal{L}_{inter} = L_{ent}(v_{c}, t_{c}) + L_{ent}(V_{c}, T_{c})$. Within each modality, we further impose parent-child structural constraints to preserve the inherent tree-like organization of surgical workflows, formulated as $\mathcal{L}_{intra} = L_{ent}(t_{c}, T_{c}) + L_{ent}(v_{c}, V_{c})$. Together, these objectives enforce hierarchical consistency both across and within modalities. 
\\
\\
\noindent \textbf{Training Paradigm.}
We adopt a two-stage optimization strategy to progressively stabilize hierarchical alignment in hyperbolic space. In \textit{Stage I}, all model parameters are optimized using only the proposed geometry-aware hyperbolic contrastive loss, aiming to establish robust cross-modal alignment while mitigating structural false negatives under hierarchical semantics. 
In \textit{Stage II}, we freeze the visual and textual encoders and fine-tune only the hyperbolic mapping layers. In this stage, the model is jointly optimized with the geometry-aware hyperbolic contrastive loss and the cone-based hyperbolic entailment losses (both inter- and intra-modal), explicitly enforcing hierarchical containment constraints. 

\section{Experiments}
\subsection{Experimental Details}
\noindent \textbf{Training Dataset.}
Following prior work~\cite{yuan2023SurgVLP}, we utilize the Surgical Video Lectures (SVL) dataset shared by the authors of ~\cite{yuan2023SurgVLP} and reorganize it into an explicit hierarchical video-text
corpus for structured pretraining. SVL contains large-scale surgical lecture videos with ASR-generated transcripts and structured annotations such as keystep descriptions and video-level abstracts~\cite{Yuan2024HecVL}. Instead of treating video-text pairs as flat clip-caption alignments, we explicitly construct parent-child units at three levels: fine-grained clips with local descriptions, phase-level segments with procedural summaries, and full videos with global abstracts. Parent segments are temporally composed of child clips, forming a tree-structured workflow hierarchy. This design enables supervision across modalities and across semantic levels through explicit containment relations. 
\\
\\
\noindent \textbf{Downstream Datasets and Evaluation.}
To validate the generalization capability of our method, we evaluate on three established surgical benchmarks: Cholec80~\cite{twinanda2016endonetcholec80} (cholecystectomy), AutoLaparo~\cite{wang2022autolaparo} (hysterectomy), and Multibypass140~\cite{lavanchy2024challengesmultibypass} (gastric bypass). Multibypass140 contains 140 laparoscopic Roux-en-Y gastric bypass surgeries annotated with 12 phases and 46 steps. Following prior work, we report results separately on its two center-specific subsets, StrasBypass70 (Stras70) and BernBypass70 (Bern70), each comprising 70 cases collected from the University Hospital of Strasbourg and Bern University Hospital, respectively, to assess cross-center generalization. All these datasets differ from our training data in both surgical type and medical center, introducing substantial domain shifts. Performance is evaluated on the official test splits. In all tables, results are reported as Top-1 accuracy and F1-score (Acc/F1).
\\
\\
\noindent \textbf{Implementation Details.}
We adopt ResNet-50~\cite{he2016resnet} as the visual encoder $\mathcal{F}_v$ and BioClinicalBERT~\cite{huang2019clinicalbert} as the textual encoder $\mathcal{F}_t$, both initialized from pretrained weights. In Stage I, training is performed with a global batch size of 256 on 8 NVIDIA A100 GPUs using the AdamW optimizer, with an initial learning rate of $5\times10^{-5}$. In Stage II, training is conducted on a single GPU with batch size 128 and a reduced learning rate of $1\times10^{-5}$. For fair comparison, all baselines follow the same SVL dataset and protocol as PeskaVLP~\cite{peskavlp}. 
\subsection{Results}

\begin{table}[t]
\centering
\caption{Zero-shot Phase Recognition across Procedures and Centers.}~\label{tab:zs}
\begin{tabular}{lccccc}
\toprule
Model & Dataset & Cholec80 & AutoLaparo & Stras70 & Bern70 \\
\midrule
MIL-NCE &  HowTo100M  & 7.8 / 7.3 & 9.9 / 7.9 & 5.6 / 3.1 & 2.4 / 2.1 \\
\midrule
\multirow{3}{*}{CLIP}
& CLIP400M & 30.8 / 13.1 & 17.4 / 9.1 & 16.9 / 5.5 & 14.8 / 4.1 \\
& Scratch  & 29.4 / 10.4 & 15.3 / 10.9 & 6.3 / 3.5  & 4.9 / 2.3 \\
& SVL & 33.8 / 19.6 & 18.9 / 16.2 & 15.8 / 8.6 & 17.8 / 7.1 \\
\midrule
SurgVLP & SVL & 34.7 / 24.4 & 21.3 / 16.6 & 10.8 / 6.9 & 11.4 / 7.2 \\
HecVL & SVL & 41.7 / 26.3 & 23.3 / 18.9 & 26.9 / 18.3 & 22.8 / 13.6 \\
PeskaVLP & SVL & 45.1 / 34.2 & 26.5 / 23.6 & 46.7 / 28.6 & 45.7 / 22.6 \\
HyperVLP & SVL & \textbf{ 49.9 / 37.2  } & \textbf{ 42.9 / 32.9  } & \textbf{ 52.7 / 41.1  } & \textbf{ 50.0 / 30.3} \\
\bottomrule
\end{tabular}
\end{table}

\begin{table*}[t]
\centering
\caption{Linear-probing evaluation results. V: supervision is from visual frames. L: supervision is from natural languages. VL: supervision is from both visual and language entities.}
\label{tab:linear_probing}
\begin{tabular}{llcccccc}
\toprule
Model & Dataset & shot & Cholec80 & Autolaparo & Stras70 & Bern70 \\
\midrule

\multirow{2}{*}{ImageNet} & \multirow{2}{*}{ImageNet} 
& 100 & 66.4 / 54.9 & 57.5 / 44.9 & 66.2 / 53.6 & 64.7 / 31.6 \\
& & 10 & 57.4 / 42.3 & 44.9 / 30.4 & 53.3 / 42.1 & 53.3 / 25.6 \\
\midrule

\multirow{2}{*}{Moco~\cite{He_2020_moco}} & \multirow{2}{*}{SVL(V)}
& 100 & 68.2 / 55.8 & 59.5 / 48.4 & 71.6 / 58.1 & 69.6 / 36.5 \\
& & 10 & 57.6 / 43.5 & 49.9 / 34.6 & 63.1 / 49.3 & 59.1 / 29.9 \\
\midrule

\multirow{2}{*}{Moco~\cite{He_2020_moco}} & \multirow{2}{*}{Cholec80}
& 100 & 73.4 / 62.8 & 51.3 / 37.4 & 67.8 / 55.4 & 66.0 / 33.1 \\
& & 10 & 69.6 / 56.9 & 45.4 / 31.7 & 58.1 / 45.2 & 52.7 / 25.7 \\
\midrule

\multirow{2}{*}{CLIP} & \multirow{2}{*}{NA(L)}
& 100 & 64.8 / 50.7 & 58.5 / 46.1 & 65.4 / 50.6 & 64.1 / 33.3 \\
& & 10 & 57.5 / 40.0 & 46.2 / 31.4 & 54.3 / 42.1 & 52.8 / 27.9 \\
\midrule

\multirow{2}{*}{CLIP} & \multirow{2}{*}{SVL(L)}
& 100 & 64.9 / 55.0 & 53.1 / 42.1 & 69.1 / 55.7 & 68.2 / 35.2 \\
& & 10 & 58.9 / 42.3 & 45.3 / 35.3 & 58.2 / 45.2 & 56.5 / 29.8 \\
\midrule

\multirow{2}{*}{SurgVLP} & \multirow{2}{*}{SVL(L)}
& 100 & 63.5 / 50.3 & 54.3 / 41.8 & 65.8 / 50.0 & 66.5 / 34.3 \\
& & 10 & 55.0 / 39.9 & 48.5 / 32.0 & 57.0 / 44.0 & 57.7 / 28.5 \\
\midrule

\multirow{2}{*}{HecVL} & \multirow{2}{*}{SVL(L)}
& 100 & 66.0 / 53.2 & 56.9 / 44.2 & 69.8 / 54.9 & 70.0 / 34.4 \\
& & 10 & 56.1 / 40.3 & 46.9 / 32.1 & 60.2 / 46.8 & 59.3 / 31.2 \\
\midrule

\multirow{2}{*}{PeskaVLP} & \multirow{2}{*}{SVL(VL)}
& 100 & 69.9 / 59.8 & 63.1 / 49.7 & 71.4 / 59.5 & 71.5 / 37.4 \\
& & 10 & 61.9 / 50.6 & 53.1 / 36.8 & 63.8 / 50.4 & 62.9 / 32.7 \\
\midrule

\multirow{2}{*}{HyperVLP} & \multirow{2}{*}{SVL(VL)}
& 100 & \textbf{ 71.3 / 60.2 } & \textbf{65.3 / 51.1} & \textbf{ 75.6 / 62.3 } & \textbf{ 76.9 / 45.1 } \\
& & 10 & \textbf{ 64.2 / 51.3 } & \textbf{ 60.1 /43.8 } & \textbf{ 66.9 / 53.4 } & \textbf{ 72.3 / 44.8} \\
\bottomrule
\end{tabular}
\end{table*}

\begin{figure}[htbp]
    \centering
    \begin{subfigure}[b]{0.48\textwidth}
        \centering
\includegraphics[width=\textwidth]{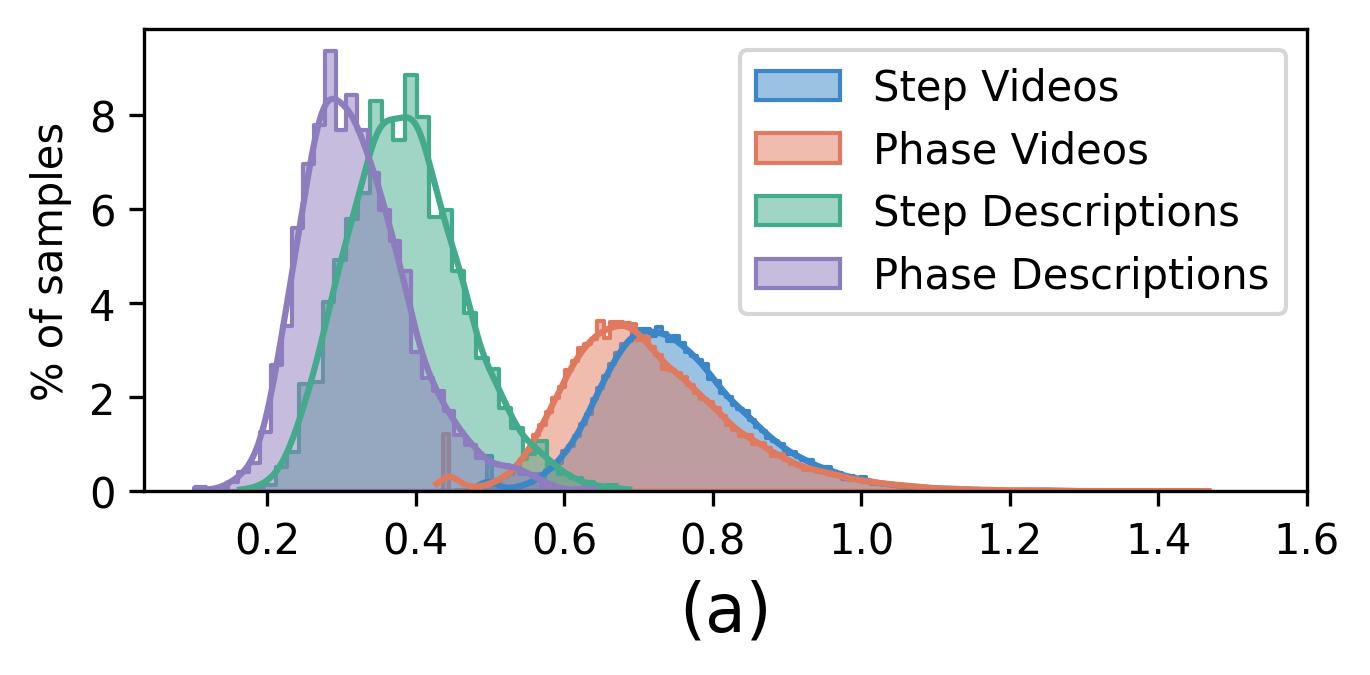}
    \end{subfigure}
    \hfill 
    \begin{subfigure}[b]{0.48\textwidth}
        \centering
\includegraphics[width=\textwidth]{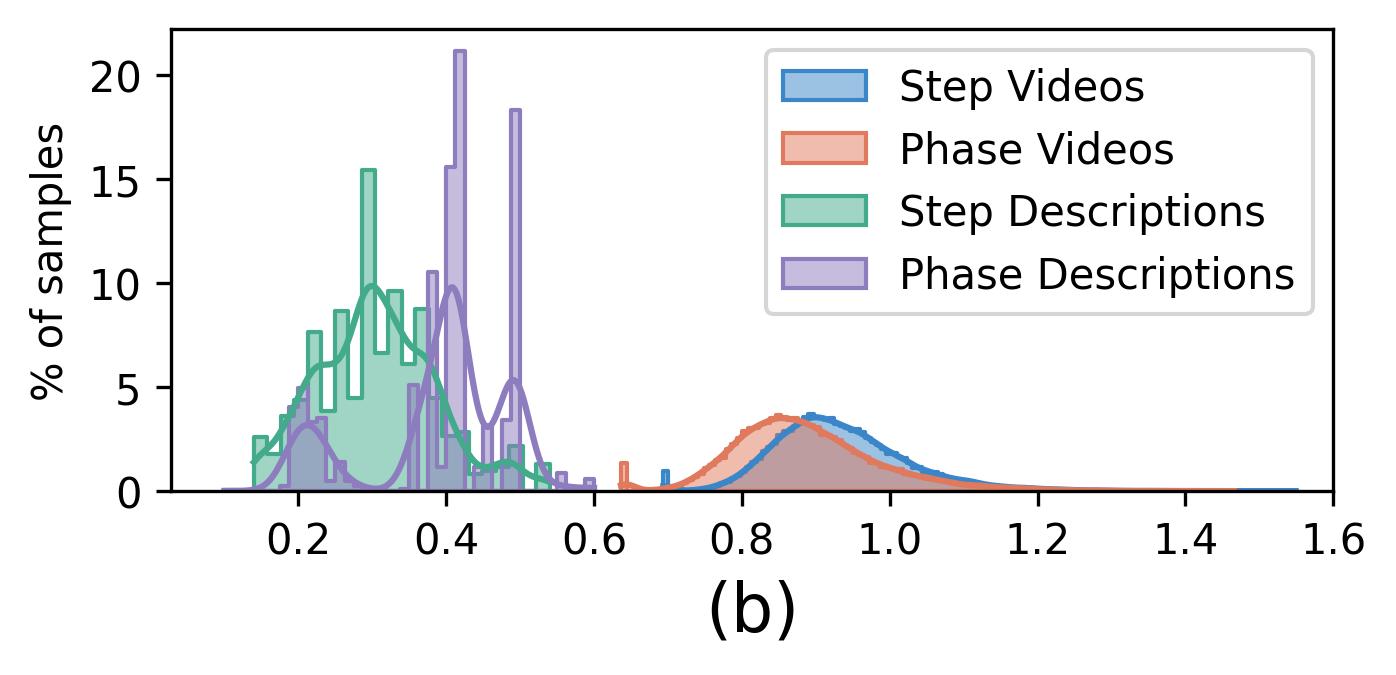} 
    \end{subfigure}
    \caption{Distribution of hyperbolic distances to the origin for video and text representations.
The x-axis shows the hyperbolic radial distance of each embedding.
Results are shown for Strasbypass70 (a) and BernBypass70 (b), including step-/phase-level videos and their corresponding descriptions.}
\label{fig:hyperbolic_norm_dist}
\end{figure}

\noindent \textbf{Zero-shot Phase Recognition.}
Table~\ref{tab:zs} reports zero-shot phase recognition across procedures (Cholec80, AutoLaparo) and centers (Stras70, Bern70). 
General-purpose models (MIL-NCE~\cite{miech2019howto100m}, CLIP~\cite{radford2021clip}) show limited transferability, especially under cross-procedure and cross-center settings. 
Surgical VLMs pretrained on SVL substantially improve performance, confirming the benefit of domain-specific pretraining. 
Among all methods, \model consistently achieves the best results on all four benchmarks, with clear margins on both Accuracy and F1-score, particularly under cross-center evaluation (Stras70 and Bern70). 
Notably, HecVL and PeskaVLP are themselves Euclidean \emph{hierarchical} surgical VLP methods built on the same SVL data and backbone. The consistent improvement of \model over them isolates the benefit of modeling the workflow hierarchy in hyperbolic rather than Euclidean space, beyond non-hierarchical CLIP/SurgVLP-style baselines.
\\
\\
\noindent \textbf{Few-/Full shot Linear Probing.}
Following the protocol of PeskaVLP~\cite{peskavlp}, we evaluate \model under 10\% and 100\% settings with frozen encoders and a linear classifier using V, L, and VL supervision. As shown in Table~\ref{tab:linear_probing}, general-purpose models show limited cross-procedure transfer, especially in the few-shot regime. Surgical pretraining improves performance, while \model consistently achieves the best results across datasets and supervision settings. The pronounced gains in the 10-shot setting indicate that our hyperbolic hierarchical modeling yields more linearly separable and procedure-agnostic representations.
\\
\\
\noindent \textbf{Qualitative Analysis.}
To analyze how different semantic granularities are organized in hyperbolic space, we examine the distribution of embedding distances to the origin, where the radial coordinate reflects the level of hierarchical abstraction. Figure~\ref{fig:hyperbolic_norm_dist} presents the radial distance distributions of step- and phase-level video embeddings, along with their paired textual representations. The textual descriptions are generated with the assistance of a large language model~\cite{hurst2024gpt4o} to ensure semantic consistency.
We observe a clear separation between semantic levels as well as between modalities. Importantly, the relative ordering of step- and phase-level representations remains consistent across StrasBypass70 (a) and BernBypass70 (b), indicating that the learned hyperbolic space preserves a stable and procedure-invariant hierarchical structure.
\\
\\
\noindent \textbf{Ablation Study.}
We evaluate the contribution of each loss component to phase recognition.  
As shown in Table~\ref{tab:ablation}, adding the geometry-aware contrastive loss ($\mathcal{L}_{GeoHCL}$), inter-modal entailment ($\mathcal{L}_{inter}$), and intra-modal entailment ($\mathcal{L}_{intra}$) progressively improves performance across datasets. 
$\mathcal{L}_{GeoHCL}$ and $\mathcal{L}_{inter}$ provide the primary gains, while $\mathcal{L}_{intra}$ further enhances hierarchical consistency. 
The best results are achieved when all components are jointly optimized.

\begin{table}[t]
    \caption{Ablation study.
    A checkmark (\checkmark) indicates the model trained with the loss function, while a dash (``-'') indicates the model trained without it. }
    \label{tab:ablation}
    \centering
\begin{tabular}{ccccccc}
\toprule
$\mathcal{L}_{GeoHCL}\quad $ & $\mathcal{L}_{inter } $& $\mathcal{L}_{intra}$ 
& Cholec80 & AutoLaparo & Stras70 & Bern70 \\
\midrule
- & - & - 
& 33.8 / 19.6 & 18.9 / 16.2 & 15.8 / 8.6 & 17.8 / 7.1 \\
\checkmark & - & - 
& 43.1 / 28.2 & 30.3 / 24.2 & 45.7 / 27.6 & 36.9 / 21.7 \\
\checkmark & \checkmark & - 
& 45.1 / 27.2 & 39.8 / 27.2 & 47.5 / 38.3 & 46.5 / 26.4 \\
\checkmark & - & \checkmark 
& 47.6 / 29.8 & 40.9 / 29.6 & 47.3 / 37.3 & 44.1 / 28.8 \\
\rowcolor{Ash grey!10}
\checkmark & \checkmark & \checkmark 
& \textbf{ 49.9 / 37.2 } 
& \textbf{ 42.9 / 32.9 } 
& \textbf{ 52.7 / 41.1 } 
& \textbf{ 50.0 / 30.3} \\
\bottomrule

\end{tabular}
\end{table}

\section{Conclusion}
In this paper, we presented \model, a hyperbolic surgical video-language pretraining framework that explicitly models the intrinsic tree structure of surgical workflows. Unlike prior Euclidean approaches that treat workflow labels as flat categories, our method embeds visual and textual representations into a shared Lorentz manifold to preserve hierarchical containment. We introduced a geometry-aware hyperbolic contrastive objective to alleviate structural false negatives and a cone-based entailment mechanism to enforce parent-child consistency across and within modalities. A progressive two-stage optimization strategy further stabilizes hierarchical alignment in hyperbolic space. Extensive experiments on three publicly available benchmarks demonstrate consistent improvements in zero-shot and few-shot phase recognition, particularly under cross-procedure and cross-center domain shifts. 

\section{Acknowledgments}
This research was partially supported by "Pioneer" and "Leading Goose" R\&D Program of Zhejiang under Grant No. 2025C01132. This work benefited from French state funds managed by the National Research Agency via the IHU Strasbourg (ANR-10-IAHU-0002) and the ENACT AI Cluster (ANR-23- IACL-0004).
\bibliographystyle{splncs04}
\bibliography{reference}

@article{maier2022surgical_background,
  title={Surgical data science--from concepts toward clinical translation},
  author={Maier-Hein, Lena and Eisenmann, Matthias and Sarikaya, Duygu and M{\"a}rz, Keno and Collins, Toby and Malpani, Anand and Fallert, Johannes and Feussner, Hubertus and Giannarou, Stamatia and Mascagni, Pietro and others},
  journal={Medical image analysis},
  volume={76},
  pages={102306},
  year={2022},
  publisher={Elsevier}
}

@inproceedings{miech2019howto100m,
  title={Howto100m: Learning a text-video embedding by watching hundred million narrated video clips},
  author={Miech, Antoine and Zhukov, Dimitri and Alayrac, Jean-Baptiste and Tapaswi, Makarand and Laptev, Ivan and Sivic, Josef},
  booktitle={Proceedings of the IEEE/CVF international conference on computer vision},
  pages={2630--2640},
  year={2019}
}

@article{twinanda2016endonetcholec80,
  title={Endonet: a deep architecture for recognition tasks on laparoscopic videos},
  author={Twinanda, Andru P and Shehata, Sherif and Mutter, Didier and Marescaux, Jacques and De Mathelin, Michel and Padoy, Nicolas},
  journal={IEEE transactions on medical imaging},
  volume={36},
  number={1},
  pages={86--97},
  year={2016},
  publisher={IEEE}
}

@inproceedings{wang2022autolaparo,
  title={Autolaparo: A new dataset of integrated multi-tasks for image-guided surgical automation in laparoscopic hysterectomy},
  author={Wang, Ziyi and Lu, Bo and Long, Yonghao and Zhong, Fangxun and Cheung, Tak-Hong and Dou, Qi and Liu, Yunhui},
  booktitle={International Conference on Medical Image Computing and Computer-Assisted Intervention},
  pages={486--496},
  year={2022},
  organization={Springer}
}

@article{peskavlp,
  title={Procedure-aware surgical video-language pretraining with hierarchical knowledge augmentation},
  author={Yuan, Kun and Navab, Nassir and Padoy, Nicolas and others},
  journal={Advances in Neural Information Processing Systems},
  volume={37},
  pages={122952--122983},
  year={2024}
}

@article{yuan2023SurgVLP,
  title={Learning multi-modal representations by watching hundreds of surgical video lectures},
  author={Yuan, Kun and Srivastav, Vinkle and Yu, Tong and Lavanchy, Joel L and Marescaux, Jacques and Mascagni, Pietro and Navab, Nassir and Padoy, Nicolas},
  journal={Medical Image Analysis},
  volume={105},
  pages={103644},
  year={2025},
  publisher={Elsevier}
}

@inproceedings{Yuan2024HecVL,
  title={Hecvl: Hierarchical video-language pretraining for zero-shot surgical phase recognition},
  author={Yuan, Kun and Srivastav, Vinkle and Navab, Nassir and Padoy, Nicolas},
  booktitle={International Conference on Medical Image Computing and Computer-Assisted Intervention},
  pages={306--316},
  year={2024},
  organization={Springer}
}

@InProceedings{Yan_Surgformer_MICCAI2024,
        author = { Yang, Shu and Luo, Luyang and Wang, Qiong and Chen, Hao},
        title = { { Surgformer: Surgical Transformer with Hierarchical Temporal Attention for Surgical Phase Recognition } },
        booktitle = {proceedings of Medical Image Computing and Computer Assisted Intervention -- MICCAI 2024},
        year = {2024},
        publisher = {Springer Nature Switzerland},
        volume = {LNCS 15006},
        month = {October},
        page = {606 -- 616}
}

@InProceedings{Pér_MuST_MICCAI2024,
        author = { Pérez, Alejandra and Rodríguez, Santiago and Ayobi, Nicolás and Aparicio, Nicolás and Dessevres, Eugénie and Arbeláez, Pablo},
        title = { { MuST: Multi-Scale Transformers for Surgical Phase Recognition } },
        booktitle = {proceedings of Medical Image Computing and Computer Assisted Intervention -- MICCAI 2024},
        year = {2024},
        publisher = {Springer Nature Switzerland},
        volume = {LNCS 15006},
        month = {October},
        page = {422 -- 432}
}

@inproceedings{radford2021clip,
  title={Learning transferable visual models from natural language supervision},
  author={Radford, Alec and Kim, Jong Wook and Hallacy, Chris and Ramesh, Aditya and Goh, Gabriel and Agarwal, Sandhini and Sastry, Girish and Askell, Amanda and Mishkin, Pamela and Clark, Jack and others},
  booktitle={International conference on machine learning},
  pages={8748--8763},
  year={2021},
  organization={PMLR}
}

@inproceedings{li2022blip,
  title={Blip: Bootstrapping language-image pre-training for unified vision-language understanding and generation},
  author={Li, Junnan and Li, Dongxu and Xiong, Caiming and Hoi, Steven},
  booktitle={International conference on machine learning},
  pages={12888--12900},
  year={2022},
  organization={PMLR}
}

@inproceedings{Desai2023meru,
author = {Desai, Karan and Nickel, Maximilian and Rajpurohit, Tanmay and Johnson, Justin and Vedantam, Ramakrishna},
title = {Hyperbolic image-text representations},
year = {2023},
publisher = {JMLR.org},
booktitle = {Proceedings of the 40th International Conference on Machine Learning},
articleno = {305},
numpages = {38},
location = {Honolulu, Hawaii, USA},
series = {ICML'23}
}

@article{PalSDFGM2024hypercoclip,
  title={Compositional entailment learning for hyperbolic vision-language models},
  author={Pal, Avik and Van Spengler, Max and di Melendugno, Guido Maria D'Amely and Flaborea, Alessandro and Galasso, Fabio and Mettes, Pascal},
  journal={arXiv preprint arXiv:2410.06912},
  year={2024}
}

@inproceedings{ganea2018hyperboliccones,
  title={Hyperbolic entailment cones for learning hierarchical embeddings},
  author={Ganea, Octavian and B{\'e}cigneul, Gary and Hofmann, Thomas},
  booktitle={International conference on machine learning},
  pages={1646--1655},
  year={2018},
  organization={PMLR}
}

@InProceedings{pmc-clip,
author="Lin, Weixiong
and Zhao, Ziheng
and Zhang, Xiaoman
and Wu, Chaoyi
and Zhang, Ya
and Wang, Yanfeng
and Xie, Weidi",
editor="Greenspan, Hayit
and Madabhushi, Anant
and Mousavi, Parvin
and Salcudean, Septimiu
and Duncan, James
and Syeda-Mahmood, Tanveer
and Taylor, Russell",
title="PMC-CLIP: Contrastive Language-Image Pre-training Using Biomedical Documents",
booktitle="Medical Image Computing and Computer Assisted Intervention -- MICCAI 2023",
year="2023",
publisher="Springer Nature Switzerland",
address="Cham",
pages="525--536",
isbn="978-3-031-43993-3"
}

@InProceedings{He_2020_moco,
author = {He, Kaiming and Fan, Haoqi and Wu, Yuxin and Xie, Saining and Girshick, Ross},
title = {Momentum Contrast for Unsupervised Visual Representation Learning},
booktitle = {Proceedings of the IEEE/CVF Conference on Computer Vision and Pattern Recognition (CVPR)},
month = {June},
year = {2020}
}

@inproceedings{he2025endo,
  title={Endo-CLIP: Progressive Self-supervised Pre-training on Raw Colonoscopy Records},
  author={He, Yili and Zhu, Yan and Fu, Peiyao and Yang, Ruijie and Chen, Tianyi and Wang, Zhihua and Li, Quanlin and Zhou, Pinghong and Yang, Xian and Wang, Shuo},
  booktitle={International Conference on Medical Image Computing and Computer-Assisted Intervention},
  pages={106--116},
  year={2025},
  organization={Springer}
}

@inproceedings{wang2023foundation,
  title={Foundation model for endoscopy video analysis via large-scale self-supervised pre-train},
  author={Wang, Zhao and Liu, Chang and Zhang, Shaoting and Dou, Qi},
  booktitle={International conference on medical image computing and computer-assisted intervention},
  pages={101--111},
  year={2023},
  organization={Springer}
}

@article{schmidgall2024general,
  title={General surgery vision transformer: A video pre-trained foundation model for general surgery},
  author={Schmidgall, Samuel and Kim, Ji Woong and Jopling, Jeffrey and Krieger, Axel},
  journal={arXiv preprint arXiv:2403.05949},
  year={2024}
}

@inproceedings{le-etal-2019-inferring,
    title = "Inferring Concept Hierarchies from Text Corpora via Hyperbolic Embeddings",
    author = "Le, Matthew  and
      Roller, Stephen  and
      Papaxanthos, Laetitia  and
      Kiela, Douwe  and
      Nickel, Maximilian",
    editor = "Korhonen, Anna  and
      Traum, David  and
      M{\`a}rquez, Llu{\'i}s",
    booktitle = "Proceedings of the 57th Annual Meeting of the Association for Computational Linguistics",
    month = jul,
    year = "2019",
    address = "Florence, Italy",
    publisher = "Association for Computational Linguistics",
    url = "https://aclanthology.org/P19-1313/",
    doi = "10.18653/v1/P19-1313",
    pages = "3231--3241",
}

@article{cannon1997hyperbolicgeometry,
  title={Hyperbolic geometry},
  author={Cannon, James W and Floyd, William J and Kenyon, Richard and Parry, Walter R and others},
  journal={Flavors of geometry},
  volume={31},
  number={59-115},
  pages={2},
  year={1997},
  publisher={Citeseer}
}

@article{lavanchy2024challengesmultibypass,
  title={Challenges in multi-centric generalization: phase and step recognition in Roux-en-Y gastric bypass surgery},
  author={Lavanchy, Jo{\"e}l L and Ramesh, Sanat and Dall’Alba, Diego and Gonzalez, Cristians and Fiorini, Paolo and M{\"u}ller-Stich, Beat P and Nett, Philipp C and Marescaux, Jacques and Mutter, Didier and Padoy, Nicolas},
  journal={International journal of computer assisted radiology and surgery},
  pages={1--9},
  year={2024},
  publisher={Springer}
}

@article{oord2018cpc,
  title={Representation learning with contrastive predictive coding},
  author={Oord, Aaron van den and Li, Yazhe and Vinyals, Oriol},
  journal={arXiv preprint arXiv:1807.03748},
  year={2018}
}

@inproceedings{he2016resnet,
  title={Deep residual learning for image recognition},
  author={He, Kaiming and Zhang, Xiangyu and Ren, Shaoqing and Sun, Jian},
  booktitle={Proceedings of the IEEE conference on computer vision and pattern recognition},
  pages={770--778},
  year={2016}
}

@article{huang2019clinicalbert,
  title={Clinicalbert: Modeling clinical notes and predicting hospital readmission},
  author={Huang, Kexin and Altosaar, Jaan and Ranganath, Rajesh},
  journal={arXiv preprint arXiv:1904.05342},
  year={2019}
}

@article{liu2025survey,
  title={A Survey of Embodied AI in Healthcare: Techniques, Applications, and Opportunities},
  author={Liu, Yihao and Cao, Xu and Chen, Tingting and Jiang, Yankai and You, Junjie and Wu, Minghua and Wang, Xiaosong and Feng, Mengling and Jin, Yaochu and Chen, Jintai},
  journal={arXiv preprint arXiv:2501.07468},
  year={2025}
}

@article{hurst2024gpt4o,
  title={Gpt-4o system card},
  author={Hurst, Aaron and Lerer, Adam and Goucher, Adam P and Perelman, Adam and Ramesh, Aditya and Clark, Aidan and Ostrow, AJ and Welihinda, Akila and Hayes, Alan and Radford, Alec and others},
  journal={arXiv preprint arXiv:2410.21276},
  year={2024}
}

@article{perez2025surglavi,
  title={SurgLaVi: Large-Scale Hierarchical Dataset for Surgical Vision--Language Representation Learning},
  author={Perez, Alejandra and Nwoye, Chinedu and Kermani, Ramtin Raji and Mohareri, Omid and Jamal, Muhammad Abdullah},
  journal={Medical Image Analysis},
  pages={103982},
  year={2026},
  publisher={Elsevier}
}

@inproceedings{wang2022medclip,
  title={Medclip: Contrastive learning from unpaired medical images and text},
  author={Wang, Zifeng and Wu, Zhenbang and Agarwal, Dinesh and Sun, Jimeng},
  booktitle={Proceedings of the 2022 Conference on Empirical Methods in Natural Language Processing},
  pages={3876--3887},
  year={2022}
}

\end{document}